%% file: main.tex
\documentclass[11pt]{article}

\usepackage[final]{acl}
\usepackage{enumitem}
\usepackage{times}
\usepackage{latexsym}
\usepackage{comment}
\usepackage{booktabs} 
\usepackage{tcolorbox} 
\tcbuselibrary{skins}
\usepackage[T1]{fontenc}

\usepackage[utf8]{inputenc}

\usepackage{microtype}
\usepackage[hang,flushmargin]{footmisc} 

\usepackage{inconsolata}
\usepackage[disable]{todonotes}  
\usepackage{graphicx}
\graphicspath{{figures/}}
\usepackage{float}
\usepackage[table]{xcolor}
\usepackage{booktabs}
\usepackage{tabularx} 
\usepackage{colortbl} 
\usepackage{xcolor}   
\usepackage{colortbl} 

\usepackage{xcolor}
\usepackage{tcolorbox}

\newcommand{\userstudyhead}[1]{\par\smallskip\noindent\textbf{#1.}\ }

\newenvironment{compactstudyquote}
{\begin{tcolorbox}[
  colback=gray!10,
  colframe=gray!45,
  boxrule=0pt,
  leftrule=2.5pt,
  sharp corners,
  boxsep=0pt,
  left=6pt,
  right=6pt,
  top=4pt,
  bottom=4pt,
  before skip=6pt,
  after skip=7pt,
  width=\linewidth,
  before upper={\normalsize\itshape\setlength{\parindent}{0pt}}
]}
{\end{tcolorbox}}

\title{\textsc{WikiStar}: A System for Shedding Light on the Hidden History\\of Scientific Wikipedia Articles}

\author{
Omer~Ehrlich$^{1*}$ \quad
Nitzan~Barzilay$^{1*}$ \quad
Rona~Aviram$^{3}$ \quad
Tom~Hope$^{1,2}$ \\
$^1$ The Hebrew University of Jerusalem \quad $^2$ Allen Institute for AI (Ai2) \\ $^3$ Department of Communication Studies, Ben-Gurion University of the Negev \\
\texttt{omer.ehrlich@mail.huji.ac.il, nitzan.barzilay@mail.huji.ac.il}
}

\newcommand{\coderepo}{\href{https://github.com/omerehrlich/WikiStar}{our code repository}}
\begin{document}
\pagestyle{plain}
\maketitle

\begingroup
\renewcommand\thefootnote{*}
\footnotetext{Equal contribution.}
\endgroup

\begin{abstract}
\input{sections/00-abstract}

\end{abstract}

\input{sections/01-introduction}
\input{sections/02-task}
\input{sections/03-system}
\input{sections/05-userstudy}

\input{sections/06-relatedwork}
\input{sections/07-conclusion}
\clearpage
\bibliography{custom}
\clearpage

\onecolumn
\appendix
\input{sections/08-appendix}
\end{document}

%% file: sections/00-abstract.tex
Wikipedia plays a key role in shaping public understanding of science, and its openly accessible revision history is a unique record of how scientific knowledge evolves over time. Yet scientifically meaningful revisions are obscured by the sheer volume of routine edits, leaving each article's scientific history hidden. We present \textsc{WikiStar} (\textbf{S}cientific \textbf{T}racking of \textbf{A}rticle \textbf{R}evisions), an interactive system for exploring scientifically meaningful changes across an article's revision history. Using an LLM classifier with an expert-designed multi-label taxonomy, \textsc{WikiStar} first tags edit types such as the addition of technical terms, new research findings, and changes in scientific narrative. Then, through interactive views, an article's full revision history can be traced at any granularity---from aggregate trends that reveal when and in which sections scientific content was added or refined, down to individual edits---showing how scientific knowledge develops at a scale previously impossible. In a user study, experts from three domains found that \textsc{WikiStar} surfaced new patterns and research questions and enabled previously impractical analyses.
We release our system, code and a human-annotated benchmark.\footnote{%
  \raisebox{-0.15\height}{\includegraphics[height=1em]{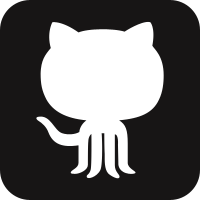}}\,%
  \href{https://github.com/omerehrlich/WikiStar}{Code}\;%
  \raisebox{-0.15\height}{\includegraphics[height=1em]{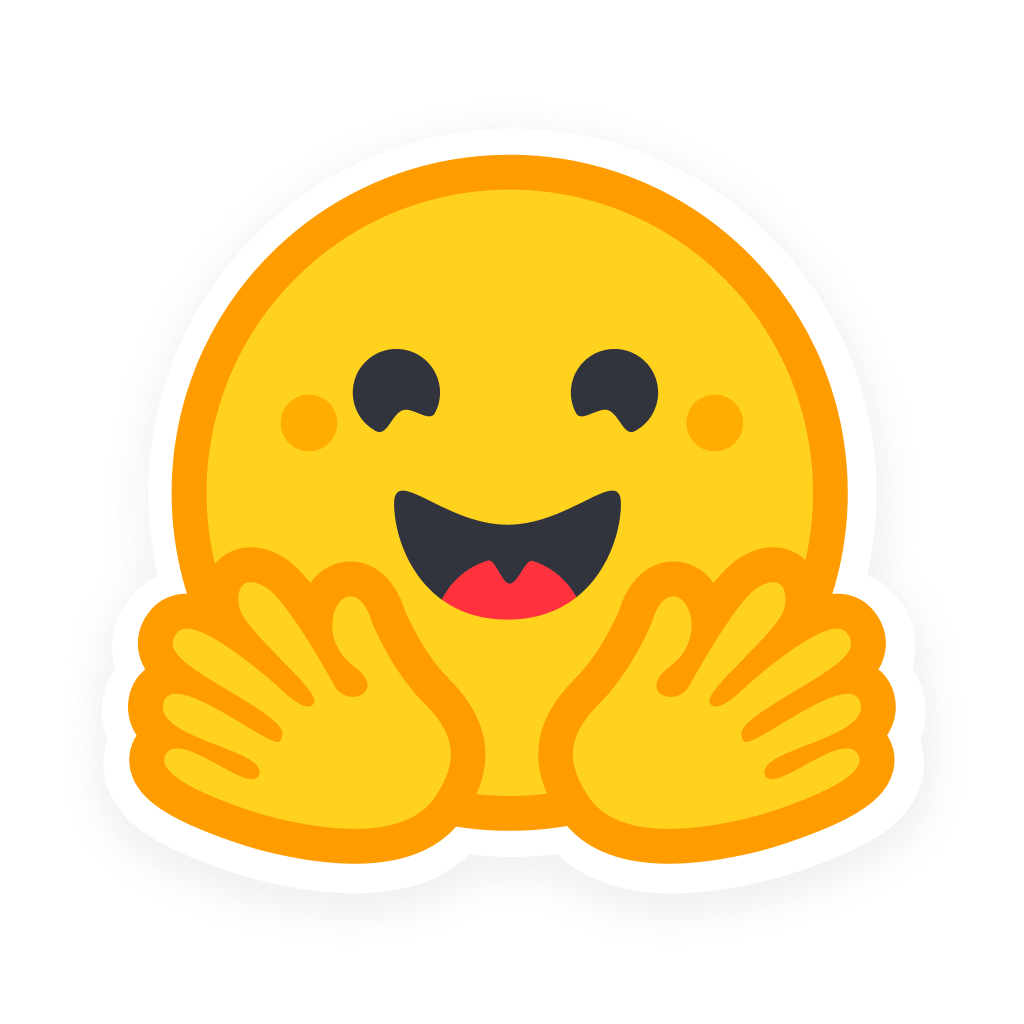}}\,%
  \href{https://huggingface.co/datasets/omerehrlich/WikiStar-Bench}{\textsc{WikiStar-Bench} dataset}\;%
  \raisebox{-0.15\height}{\includegraphics[height=1em]{img/hf-logo.png}}\,%
  \href{https://huggingface.co/spaces/omerehrlich/WikiStar}{\textsc{WikiStar} demo}}

\begin{figure}[!t]
\centering
\includegraphics[width=0.95\columnwidth]{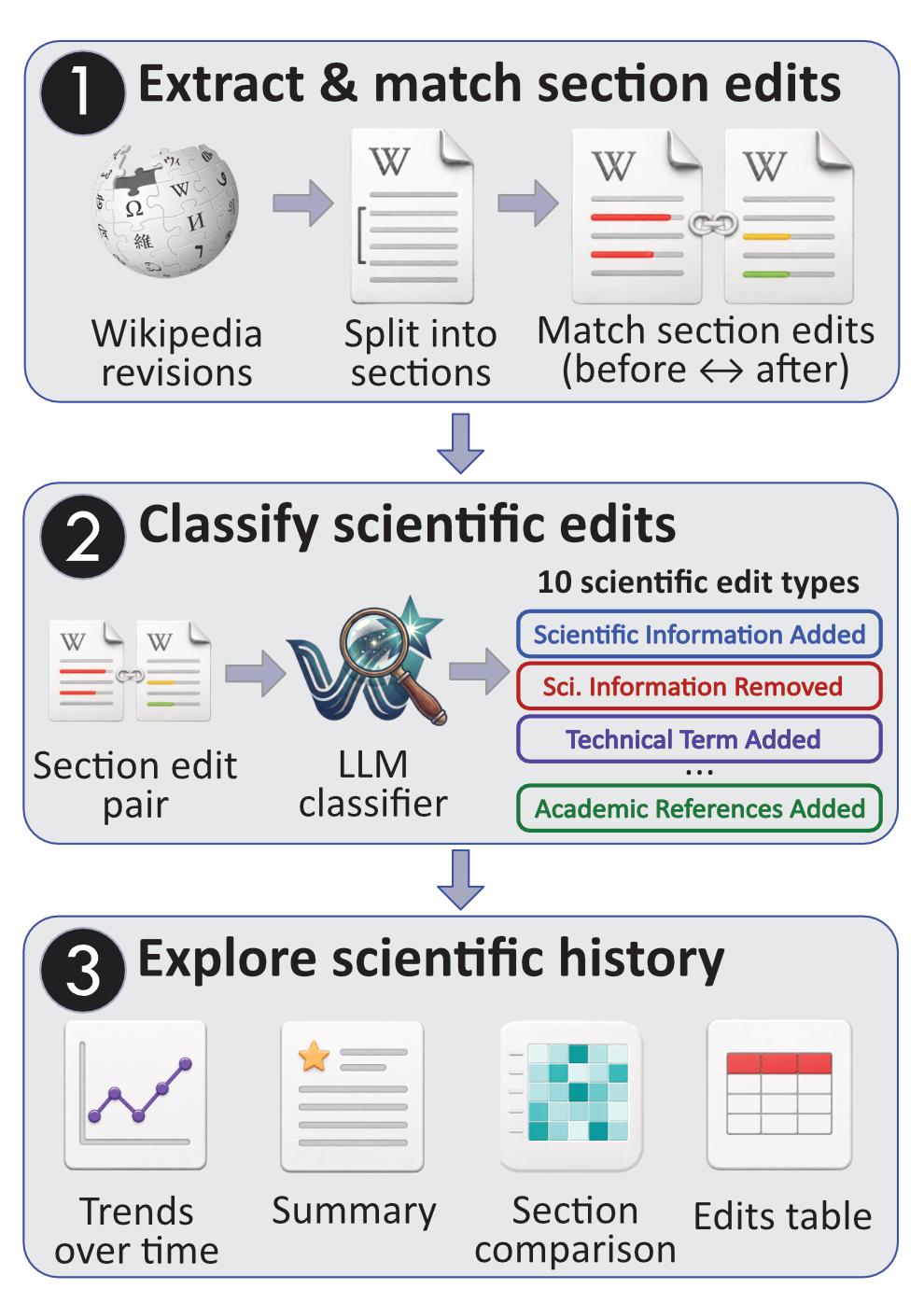}
\caption{The \textsc{WikiStar} pipeline for tracking the history of scientific edits in a Wikipedia article. (1) Splitting each revision into sections and matching them across revisions to recover the previous version of each edited section. (2) Multi-label edit-type classification of each edit pair. (3) Exploring through interactive views, from aggregate patterns down to individual edits.}
\label{fig:system-small}
\end{figure}

%% file: sections/01-introduction.tex
\section{Introduction}
\label{sec:intro}
Wikipedia, the world's largest online encyclopedia, is written and continuously rewritten by volunteers around the world. As a widely consulted source, it shapes how the public understands scientific and scholarly topics. Its openly accessible revision history lets scholars trace the development of ideas within and between academic domains, and examine how public knowledge evolves \citep{benjakob2023wikipedia, benjakob2021citation}. Scientific articles on Wikipedia are especially well suited for studying knowledge evolution. Prior work has shown that they closely track developments in the scientific literature through extensive use of reliable academic sources, making them more than public-facing summaries \citep{Simons2024WhoAT,benjakob2023wikipedia}. Their revision histories therefore potentially constitute a rich longitudinal record of how scientific concepts are introduced, debated, refined, and stabilized.

However, a single article can accumulate thousands of edits over its lifetime---the ``AI'' article alone has over 18{,}000 section edits---and scientifically significant edits make up only a small fraction, buried among routine changes such as rephrasing and formatting. Isolating meaningful edits manually is highly labor-intensive, so scientifically significant edits remain hidden and inaccessible.

Prior work classifies Wikipedia edits \citep{yang2017identifying, rajagopal2022one} and builds tools for exploring authorship and editorial conflicts \citep{viegas2004historyflow, floeck2014wikiwho, guo2023edithistory}, but both lines target general-purpose editing at the whole-page or token level. We instead work at the section level and focus on scientific significance, which localizes each edit, reveals how a section's scientific content develops over time, and drives the interactive views of our system.

We present \textbf{\textsc{WikiStar} (\textbf{S}cientific \textbf{T}racking of \textbf{A}rticle \textbf{R}evisions)}, a system that detects, quantifies, and contextualizes scientific textual changes in Wikipedia articles over time (Figure~\ref{fig:system-small}). We introduce the task of \emph{scientific edit classification}: labeling section-level edits with the scientifically meaningful changes they make, using a taxonomy of ten expert-designed labels. \textsc{WikiStar} extracts section-level edits, classifies each edit against the taxonomy using an LLM, and presents the results through interactive views that trace how an article's scientific content evolves across time and sections (Figure~\ref{fig:pipeline}). This makes the scientific edit history of Wikipedia accessible to Wikipedia researchers and more broadly to scientists, journalists, and others interested in how science is represented online. 

In a user study, experts in Wikipedia editing, science journalism, and the history and philosophy of science found that \textsc{WikiStar} surfaced patterns they could not easily discover manually, prompted new research questions, and enabled analyses that were previously impractical.

Finally, we release {\textsc{WikiStar-Bench}}, a human-annotated dataset of section-level Wikipedia edits labeled for scientific significance. 

\textbf{In summary, our main contributions:}
\begin{enumerate}[label=(\arabic*)]
    \item We create \textsc{WikiStar}, a system for exploring how a Wikipedia article's scientific content evolves over time. We validate \textsc{WikiStar} in a user study with domain experts from three domains, who found it highly useful.
    \item We introduce the task of scientific edit classification of Wikipedia article sections, grounded in our taxonomy of ten labels for scientifically salient changes, and release \textsc{WikiStar-Bench}, a human-annotated dataset of 1{,}387 section edits spanning three domains.
\end{enumerate}

%% file: sections/02-task.tex
\section{Scientific Edit Classification}
\label{sec:task}

\begin{figure*}[t]
\centering
\includegraphics[width=\textwidth,trim=20 10 20 10,clip]{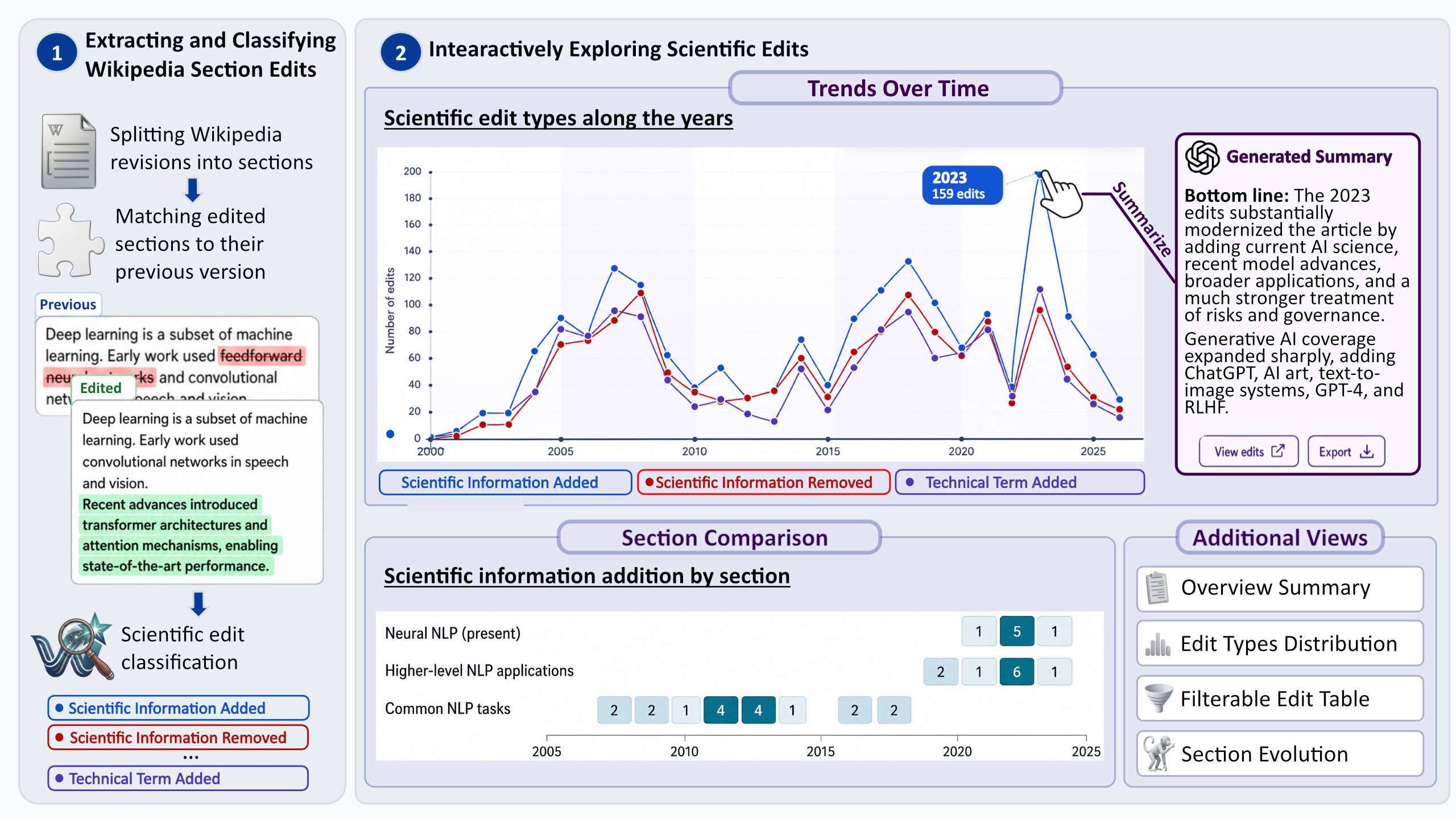}
\caption{Overview of the \textsc{WikiStar} system. \textbf{(1) Extraction and classification:} The system splits each Wikipedia revision into sections, matches each edited section to its previous version, and classifies each edit pair into scientific edit types. \textbf{(2) Interactive exploration:} Interactive views let users explore the classified history: The \textit{Over Time} view shows temporal trends in edit types, and the \textit{Section Comparison} view compares them across selected sections. Selecting a point or cell within those views generates a summary of its edits, with access to the underlying edits.}
\label{fig:pipeline}
\end{figure*}

In this section, we first introduce the task of \emph{scientific edit classification} of Wikipedia edits. To support this task, we release \textsc{WikiStar-Bench}, a human-annotated benchmark, and use it to compare seven LLMs as classifiers. 

As discussed in the Introduction, since scientifically significant edits are only a small fraction of an article's edits, we define \emph{scientific edit classification} as labeling each Wikipedia edit with the types of scientific change it makes. We cast this as a multi-label classification task over section-level edits. Given a section and its preceding version, which we call a \emph{section edit pair}, the task is to assign every applicable label from our taxonomy of ten scientific edit types (§\ref{subsec:taxonomy}), or the \textit{Non-Scientific Edit} label when no scientific change occurs.

We choose the section as our unit of granularity for two reasons. First, it carries the right amount of context for classification: whole revisions inflate prompt length and reduce output consistency (\citet{shivashankar2024contri, fieblinger2024actionable}), while individual sentences discard the context needed to judge whether a change is scientifically meaningful. Second, sections localize each edit to a specific aspect of the article, letting \textsc{WikiStar} track how scientific content is distributed across sections and develops over time.

\subsection{Section Identification and Linking}
\label{subsec:section-identification}
Turning an article's revision log into section-level edits requires two operations: parsing each revision into its constituent sections, and pairing each edited section with its previous version---or, for a newly created section, recognizing that it has none. 

\textbf{Acquiring section text}: Using the MediaWiki API \cite{mediawiki-api}, we retrieve every revision of the article as raw wikitext with its metadata (revision ID, timestamp, editor). Each revision is split into sections by a regex-based parser that follows Wikipedia's markup conventions. As each revision contains the article's full section list, most of it unchanged from the previous one, we retain only the sections that changed, so each entry corresponds to a single edited section.

\textbf{Matching section edit pairs}: The \textit{scientific edit classification} task requires the version of each section before and after an edit, which we call a \emph{section edit pair}. Matching versions is challenging because sections are not stable: editors rename, split and merge them, yet Wikipedia stores each revision as flat full-article text with no identifier linking section versions. To locate a section's earlier version, we search backward through revisions and compute two similarity measures against candidate sections: \textbf{title similarity} (normalized Levenshtein distance) and \textbf{content similarity} (cosine similarity over embeddings of the section content). We match the section to the most recent candidate for which either measure exceeds a threshold. This simple approach worked well in practice; more advanced matching is left to future work.

\subsection{Scientific Edits Taxonomy}
\label{subsec:taxonomy}
A closed set of labels is what lets \textsc{WikiStar} aggregate, filter, and visualize edits by type. We designed a taxonomy of ten scientifically meaningful edit types—such as \textit{New Scientific Information} and \textit{Change in Scientific Narrative}—plus a \textit{Non-Scientific Edit} label, built by a scientific-Wikipedia expert and refined iteratively against real edits. Full definitions appear in Appendix~\ref{sec:appendix-taxonomy}, with annotated examples for select types in Appendix~\ref{app:edit-examples}.

Some labels in the taxonomy may appear replaceable by rule-based detectors, but each requires judgment that rules do not capture. \textit{Wikilink Added} could be approximated by detecting new square-bracketed text, but such a rule would also count existing links that were only fixed or reformatted. \textit{Academic References Added} is similar: references share a common structure, but the citation format does not indicate whether the source is scientific.

\subsection{\textsc{WikiStar-Bench}}
We curate \textsc{WikiStar-Bench}, the first human-annotated dataset of scientifically labeled section-level Wikipedia edits, released as an evaluation resource for scientific edit classification. It contains 1,387 examples across three scientific domains---Biology (821), Computer Science (288), and Neuroscience (278)---each spanning at least 9 popular Wikipedia pages. Each example is a single section edit pair with its metadata and human-annotated gold labels from our taxonomy.

\subsection{Benchmark construction}
We first applied LLaMA~3.3~70B \citep{llama3} to all section-edit pairs extracted from the complete revision histories of 30 Wikipedia articles. From these model-labeled edits, we sampled examples for human annotation to obtain broad coverage across source articles and predicted edit types. The final gold-label supports are not uniform because model predictions can differ from human annotations and examples may receive multiple co-occurring labels. This sampling procedure intentionally enriches scientifically meaningful edits rather than preserving their natural prevalence. Non-Scientific Edit accounts for about 31\% of \textsc{WikiStar-Bench}, whereas most edits in complete article histories are non-scientific.

\paragraph{Annotation process:} The {WikiStar-Bench} dataset was labeled by four annotators: three STEM graduate students and an assistant professor who holds a PhD in biology and researches scientific Wikipedia. Annotators labeled each section edit pair independently, assigning every applicable label from our 10-label taxonomy, or the \textit{Non-Scientific Edit} label if none applied.

\paragraph{Annotator agreement:}
\label{sec:agreement}
To ensure reliability of human annotation, all annotators tagged a shared subset of 104 examples. 
Annotator agreement using mean pairwise Cohen's~$\kappa$
\citep{Cohen1960ACO} is high ($\kappa=0.89$), indicating strong overall agreement. Agreement is naturally lower on more subjective and interpretive labels (such as \emph{Change in Scientific Narrative} with $\kappa=0.7$ and \emph{Scientific Clarification Added} with $\kappa=0.8$). 

\subsection{Results}
\label{sec:results}
\paragraph{Experimental settings:} Against \textsc{WikiStar-Bench}'s gold labels, we report per-label precision and recall, and macro- and micro-F1 (our primary metric). We evaluate seven LLMs as classifiers: GPT-5.4, GPT-5-mini, GPT-5.1 \citep{openai2025gpt5}, GPT-4o \citep{gpt4o}, o3-mini \citep{openai2025o3mini}, LLaMA~3.3~70B \citep{llama3}, and Qwen3-Next-80B \citep{qwen3}. Scientific edits are rare in real article histories, so \textsc{WikiStar-Bench} was built to over-sample them. Because they are also the harder labels to classify, the reported scores reflect a harder setting than a full article history.

\paragraph{Prompt refinements:} Our final prompt opens with a general task explanation, then gives each label a definition and a single minimal example. The most effective refinements over a naive one-sentence-definition prompt were asking the model to first decide whether the edit is scientific at all, and flagging common mistakes in the definitions; these raise GPT-5.4's macro-F1 from $0.69$ to $0.82$. Few-shot demonstrations did not help (macro-F1 drops to $0.78$). Appendix~\ref{app:prompt-versions} reports per-model results across prompt versions.\footnote{See prompt at
\href{https://github.com/omerehrlich/WikiStar/blob/master/wiki_pipeline/prompts/create_prompt.py}{\texttt{create\_prompt.py}} in our repository.}

\begin{table}[t]
\centering
\setlength{\tabcolsep}{4pt}
\small
\begin{tabular}{@{}lrrr|r@{}}
\toprule
                                & \multicolumn{3}{c}{GPT-5.4} & Human \\
                                \cmidrule(lr){2-4}\cmidrule(lr){5-5}
Label                           & P   & R   & F1  & F1  \\
\midrule
New Sci.\ Information            & .88 & .74 & .80 & .93 \\
Sci.\ Information Removed        & .84 & .92 & .87 & .94 \\
Sci.\ Clarification Added        & .73 & .72 & .72 & .86 \\
Technical Terms Added            & .80 & .93 & .86 & .93 \\
Researcher Names Added           & .93 & .82 & .87 & .92 \\
Change in Sci.\ Narrative        & .54 & .70 & .61 & .72 \\
Academic References Added        & .90 & .86 & .88 & .97 \\
Academic References Removed      & .84 & .86 & .85 & .95 \\
Wikilink Added                   & .95 & .83 & .89 & .97 \\
Quantitative Information         & .75 & .86 & .80 & .95 \\
\cmidrule(lr){1-5}
Non-Scientific Edit              & .83 & .87 & .85 & .91 \\
\midrule
Macro avg                       & .82 & .83 & .82 & .91 \\
Micro avg                       & .84 & .83 & .83 & .92 \\
\bottomrule
\end{tabular}
\caption{Per-label precision, recall, and F1 for GPT-5.4 on \textsc{WikiStar-Bench} ($n{=}1{,}387$), and the human agreement ceiling---mean pairwise F1 across annotators on overlapping examples ($n{=}104$).}
\label{tab:perlabel_gpt54}
\end{table}

\begin{table*}[t]
\centering
\small 
\setlength{\tabcolsep}{6pt}
\begin{tabular}{@{}l cccc | cccc | cccc@{}}
\toprule
& \multicolumn{4}{c}{\shortstack{Biology\\{\scriptsize 820 examples}}}
& \multicolumn{4}{|c}{\shortstack{Computer\\Science\\{\scriptsize 288 examples}}}
& \multicolumn{4}{|c}{\shortstack{Neuroscience\\{\scriptsize 278 examples}}} \\
\cmidrule(lr){2-5}\cmidrule(lr){6-9}\cmidrule(lr){10-13}
Edit Type & P & R & F1 & \# & P & R & F1 & \# & P & R & F1 & \# \\
\midrule
New Scientific Information            & .88 & .77 & .82 & 307 & .93 & .73 & .81 & 135 & .86 & .69 & .77 & 145 \\
Scientific Information Removed         & .89 & .88 & .88 & 108 & .80 & .98 & .88 & 50  & .79 & .92 & .85 & 66  \\
Scientific Clarification Added         & .72 & .76 & .74 & 149 & .77 & .71 & .74 & 69  & .70 & .63 & .66 & 51  \\
Scientific Technical Terms Added       & .77 & .91 & .83 & 214 & .86 & .95 & .90 & 109 & .80 & .96 & .87 & 101 \\
Researcher Names Added                 & .92 & .85 & .89 & 81  & 1.00 & .64 & .78 & 28  & .90 & .86 & .88 & 42  \\
Change in Scientific Narrative         & .52 & .73 & .60 & 22  & .62 & .50 & .55 & 16  & .54 & .87 & .67 & 15  \\
Academic References Added              & .94 & .86 & .90 & 139 & .80 & .86 & .83 & 14  & .82 & .89 & .85 & 45  \\
Academic References Removed            & .87 & .85 & .86 & 47  & .83 & .83 & .83 & 6   & .79 & .89 & .84 & 26  \\
Wikilink Added                         & .95 & .84 & .89 & 234 & .96 & .86 & .91 & 109 & .95 & .78 & .85 & 117 \\
Add./Mod. of Quantitative Information & .77 & .87 & .81 & 121 & .63 & .71 & .67 & 42  & .81 & .96 & .88 & 45  \\
\cmidrule(lr){1-13}
Non Scientific Edit                       & .90 & .89 & .89 & 295 & .70 & .88 & .78 & 67  & .73 & .79 & .76 & 72  \\
\midrule
Macro Avg & .83 & .84 & .83 & 1717 & .81 & .79 & .79 & 645 & .79 & .84 & .81 & 725 \\
Micro Avg & .85 & .84 & .85 & 1717 & .83 & .81 & .82 & 645 & .81 & .82 & .82 & 725 \\
\bottomrule
\end{tabular}
\caption{\textbf{Performance Across Domains of GPT-5.4 on \textsc{WikiStar-Bench} Per-Label.} Precision (P),
Recall (R), F1, and support (\#; number of annotated instances) for
the ten scientific edit types and the \textit{Non-Scientific Edit} label.}
\label{tab:per-domain-results}
\end{table*}

\textbf{Aggregate results:} Of the seven models, GPT-5-mini and GPT-5.4 are the two best-performing, with comparable scores (macro-F1 $0.83$ vs.\ $0.82$; micro-F1 $0.86$ vs.\ $0.83$). 
The deployed demo uses GPT-5-mini due to its overall comparable results and lower operational cost. Results for all models are in Appendix~\ref{app:perlabel}. Across labels, GPT-5.4 trails the human agreement ceiling (macro-F1 $0.82$ vs.\ $0.91$), showing that scientific edit classification is not trivial and leaves room for improvement.

\paragraph{Per-Label results:} Table~\ref{tab:perlabel_gpt54} reports per-label results for GPT-5.4. F1 varies substantially by label: strongest on objective labels with explicit textual signals (such as \textit{Wikilink Added} with $0.89$, and \textit{Academic References Added} with $0.88$), where remaining errors mostly concern the scientific qualifier, e.g., counting a book as an academic reference. It is weakest on the interpretive labels (e.g., \textit{Change in Scientific Narrative} at $0.61$), where the model tends to over-extend the label beyond its intended scope, often misclassifying \textit{Scientific Clarifications} as narrative changes. This tracks lower human agreement on those labels (Section~\ref{sec:agreement}). 

\paragraph{Domain-Specific Results:} We report GPT-5.4's per-label F1 for each domain in \textsc{WikiStar-Bench}: Biology, Computer Science, and Neuroscience in Table~\ref{tab:per-domain-results}. Macro-F1 is stable across all domains. As on the full benchmark, objective labels retain high performance throughout, while interpretive labels are weakest in every domain. This consistency indicates that the taxonomy and prompts transfer across scientific domains rather than overfitting to any single one.

%% file: sections/03-system.tex
\section{\textsc{WikiStar}: System Overview}
\label{sec:system}

\textsc{WikiStar} extracts section edit pairs from an article's revision history, classifies each with our multi-label taxonomy 
(see §\ref{subsec:taxonomy}), and turns the labeled edits into interactive views that let users explore the article's scientific content across time and sections (see Figure~\ref{fig:pipeline}). Each component of \textsc{WikiStar} addresses a concrete question researchers pose about an article's scientific history: what changed, when, where, and what type of change. We introduce these components through the following scenario:

\textit{Consider a historian of science who wants to know how Wikipedia's ``Artificial Intelligence'' article has changed over time: which ideas were treated as established, when new topics entered, where editors' attention concentrated. The answer lives in the revision history, but reading it by hand is impractical, as it accumulated over 18{,}000 section edits across two decades.}

\subsection{History Overview} 

The \textit{History Overview} view is an LLM-generated summary of how the article's scientific content has evolved. Classification produces a short explanation of what changed in each edit; the \textit{Overview} gathers these explanations for every edit labeled \emph{New Scientific Information} or \emph{Change in Scientific Narrative}---the two labels capturing new or revised scientific claims---and composes them chronologically into a single structured account. A brief summary highlights the most significant changes; clicking \texttt{show more} expands it into a period-by-period narrative of the article's lifetime.

\textit{The historian begins with the History Overview: the AI article opens with broad definitions of the field, grows more technical as deep learning enters, and turns recently to societal impact---two decades of edits, visible at a glance.}

\subsection{Trends Over Time}
The \textit{Trends Over Time} view provides an interactive line plot comparing the frequency of selected edit types across time (see Figure~\ref{fig:pipeline}, \textit{Trends Over Time}).
By default, the view aggregates yearly across all sections, but supports filtering along several variables to suit each user's focus. Users can select the time frame, the temporal resolution, and the article's sections to include.
Clicking a point on the graph generates a summary of the edits it represents (those of a given edit type within a specific period), and the \textit{summarize all time periods} button extends this feature to an entire edit-type series at once; both ground the aggregated trends in the edits' textual content. Below the summary, additional options let users inspect the underlying edit text or export several types of information.

\textit{Narrowing the \textit{Trends Over Time} view to the Applications section, the historian notices a sharp rise around 2023 in new scientific information and technical terms, with few academic references. Clicking the peak, they find the summary ties the surge partly to the public release of LLMs, but also, unexpectedly, to editors choosing to add decades-old milestones like Deep Blue and Watson.}

\subsection{Per-Section Edit Heatmap}
The \textit{Section Comparison} view complements the aggregated \textit{Trends Over Time} view with fine-grained comparison across sections. For any selected edit type, it presents a heatmap where each row is a section and color intensity shows how that edit type trends within the section over time (Figure~\ref{fig:pipeline}, \textit{Section Comparison}). Like the Over Time view, it supports filtering and generates an exportable summary when a cell is clicked. Together, these enable both cross-section comparison and targeted drill-down: users can single out a section, edit type, and period, then surface the actual edits.

\textit{Switching to the \textit{Section Comparison} view, the historian sees \emph{where} each change happened, not just when: \emph{Change in Scientific Narrative} concentrates in the early 2000s in \emph{Philosophy} and \emph{History}, while \emph{New Scientific Information} surges after 2020 in \emph{Applications} and \emph{Misinformation}—a shift from debating what AI is to tracking what it does.}

\subsection{Section Evolution} 
The \textit{Section Evolution} view shows how the article's sections evolved. It lays out the sections as rows and time periods as columns, coloring each cell to show what happened to that section in that period: whether it first appeared, was edited in a scientifically meaningful way, or was renamed. This allows users to inspect the development of the structure of the article as it changed throughout the years. 

\subsection{Edit Table}
This view presents a filterable table of all edits across the article's history, together with their metadata. Users can locate specific edits by filtering on metadata fields (e.g. section names and time range) or by free-text search. Then, users see a table of section edits. Each row shows the section text before and after the edit, the edit's metadata, the scientific edit labels the model assigned along with its explanation for each assignment.To support any type of further analysis, the table can be exported.

%% file: sections/05-userstudy.tex
\section{User Study}
\label{sec:casestudies}

We conduct a user study with three participants, chosen to span complementary perspectives on how scientific knowledge is produced and evolves on Wikipedia. \textbf{P1} is deeply involved in the Wikipedia community as a writer and editor, and is a PhD candidate in Physics. \textbf{P2} is a journalist covering cybersecurity and disinformation, with an MA in the history of science. In their work, they frequently treat Wikipedia's edit record as a lens for examining how knowledge is constructed, contested, and manipulated. \textbf{P3} is a researcher in the philosophy and history of science, focused on how scientific ideas and models move between fields.

Each participant received an introduction to \textsc{WikiStar}, then had 15 minutes to freely explore a scientific Wikipedia article of their choice. They were asked to think aloud. Afterwards, they rated nine statements about the system's utility on a 5-point Likert scale (Table~\ref{tab:user-study}). The full script and participant rankings are available in \href{https://github.com/omerehrlich/WikiStar-UserStudy}{this repository}.

\begin{table}[h]
\centering
\small
\setlength{\tabcolsep}{6pt}
\renewcommand{\arraystretch}{1.15}
\begin{tabular}{@{}p{0.78\columnwidth} r@{}}
\toprule
\textbf{Statement Examples} & \textbf{Avg.} \\
\midrule
I find the types of edits \textsc{WikiStar} tracks useful and insightful. & 5.00 \\
\textsc{WikiStar} would let me do research or analysis that wasn't practical for me before. & 5.00 \\
Using \textsc{WikiStar}, I noticed potentially interesting things that I could not have easily discovered myself without it. & 4.67 \\
\textsc{WikiStar} prompted new questions about the evolution of the Wikipedia article I explored that I would not have thought to ask. & 5.00 \\
\textsc{WikiStar} could allow me to confirm or quantify things I had previously only suspected. & 4.33 \\
\textsc{WikiStar} could surface interesting evidence about how science is presented to the public. & 4.67 \\
\bottomrule
\end{tabular}
\caption{Participant agreement ratings on system-utility statements, collected via a 5-point Likert scale (1 = strongly disagree, 5 = strongly agree). \textbf{Avg.}\ denotes the mean rating across participants. A subset of the 9 rated statements is shown; see full set in \href{https://github.com/omerehrlich/WikiStar-UserStudy}{this repository}.
}
\label{tab:user-study}
\end{table}

\userstudyhead{Tracing a subject across fields}
\textbf{P3} examined the ``Ising model'' article, a physics formalism that later spread to
other fields. Puzzled that the \textit{Section Comparison} view framed its neuroscience
content as ``social sciences,'' they found the answer in \textit{Section Evolution}---the
section was renamed ``Neuroscience'' only in 2022:
\begin{compactstudyquote}
``It's very interesting to see how this comes together and explains the structure of the article.''
\end{compactstudyquote}

\userstudyhead{Surfacing invisible structural history}
\textbf{P1} used \textsc{WikiStar} to examine the ``Chaos Theory'' article, whose foundational research largely predates Wikipedia. They wanted to trace how its structure changed across twenty-five years, and while inspecting the \textit{Section Evolution} view, they noted:
\begin{compactstudyquote}
``It's kind of a dream for anyone who discusses and thinks of Wikipedia as a source of statistical data. This is much richer and more detailed than any other tool we currently have in our toolbox as Wikipedia authors and editors. This is really helpful and useful.''
\end{compactstudyquote}

\userstudyhead{From trends to edits}
\textbf{P2} used \textsc{WikiStar} to examine the ``Vaccine'' article, tracing scientific debate and controversy across its history. While looking at the \textit{Section Comparison} view, one cell sparked their interest, showing markedly more added scientific content than its neighbors. On discovering that they could generate a summary of the edits composing that cell, they enthused:
\begin{compactstudyquote}
``That's seriously gorgeous! It adds tons and tons of value. This is a very good analysis that I would actually want and need for my research. It helps me actually do my work.''
\end{compactstudyquote}

\noindent They described the experience as ``almost like guided reading.'' Turning to the \textit{Trends Over Time} view, \textbf{P2} experimented with the summarization feature across several points on the plot, then chose \textit{View source edits} to inspect the data behind the summaries and reflected:
\begin{compactstudyquote}
``Now I can connect my pipeline of thinking and research! This option to look at the metadata and the text of the edits is important, because I need to have the data itself and validate it.''
\end{compactstudyquote}

%% file: sections/06-relatedwork.tex
\section{Related Work}
\label{sec:relatedwork}
A growing body of work studies how scientific knowledge is represented and diffuses on Wikipedia \citep{Teplitskiy2015AmplifyingTI}. Most relevant to us, \citet{benjakob2023wikipedia} treat the revision history as a historiographical record of science, reading editorial revisions as intentional acts that mark how knowledge is introduced, refined, and restructured over time. Studies of this kind remain rare because they are manual and labor-intensive, requiring close reading of long revision histories. A single article accumulates thousands of revisions, most of them unrelated to scientific content, so the effort grows with the length of the history and confines such studies to a few articles.
In parallel, prior work classifies Wikipedia edits using general edit-type and intention taxonomies \citep{Liu2011WhoDW, Daxenberger2012ACS, yang2017identifying}, capturing generic operations rather than scientifically meaningful change, while another line visualizes edit provenance and activity, from token-level authorship \citep{floeck2014wikiwho} to edit wars and vandalism \citep{guo2023edithistory}. The first group labels the form of an edit, so two edits that both add a citation receive the same label whether one records a replicated result or repairs a broken link. Neither classifies the \emph{scientific} content of edits, and both operate at the page or token level rather than the section. The page level is too coarse to indicate which part of an article changed, and the token level splits a change into insertions and deletions that must be reassembled before it can be interpreted. \textsc{WikiStar} is the first to do both at the section level: it matches each edited section to its previous version and classifies the change into ten scientific types.
A separate line defines edit-intention taxonomies for revisions to scientific papers themselves \citep{Jiang2022arXivEditsUT, Ruan2024Re3AH}, tracking how the science evolves rather than how it is represented. There, revisions come from the authors and arrive in discrete versions, while Wikipedia articles are revised by many editors with different aims and no version boundary between rounds of changes. In Wikipedia, scientific edits are buried among many unrelated ones and must first be isolated, which is what \textsc{WikiStar} supports. Working at the section level lets users move between one section's history and whole-article trends, making scientific edit analysis accessible, at scale, to anyone interested in how science evolves and is represented online.

%% file: sections/07-conclusion.tex
\section{Conclusion}
\label{sec:conclusion}
We presented \textsc{WikiStar}, a system for tracking how the scientific content of Wikipedia articles develops over time. \textsc{WikiStar} extracts section-level edits, classifies each into an expert-designed taxonomy of ten scientifically meaningful edit types, and presents the results through interactive views that let users move from high-level trends down to the individual edits behind them. We released \textsc{WikiStar-Bench}, the first human-annotated benchmark for scientific edit classification, and used it to compare seven LLMs.

A user study with participants from Wikipedia editing, science journalism, and the history and philosophy of science suggests \textsc{WikiStar} is useful across these varied perspectives, enabling new kinds of research into how the representation of science on Wikipedia changes over time.

%% file: sections/08-appendix.tex
\section{Appendix - Edit-Type Taxonomy}
\label{sec:appendix-taxonomy}

\begin{table}[H]
\centering
\footnotesize
\setlength{\tabcolsep}{8pt}
\renewcommand{\arraystretch}{1.35}
\begin{tabularx}{\textwidth}{@{}>{\raggedright\arraybackslash}p{0.24\textwidth} X@{}}
\toprule
\textbf{Label} & \textbf{Definition} \\
\midrule
New Scientific Information &
Addition of information in the current revision absent from the previous one. \\
\rowcolor{gray!12}
Scientific Information Removed &
Removal of at least one complete sentence carrying scientifically meaningful content (factual claims, findings, or definitions) present before but not now; excludes vandalism reverts and non-substantive wording. \\
Scientific Clarification Added &
Modification or expansion of scientifically relevant information already present in the previous revision. \\
\rowcolor{gray!12}
Scientific Technical Terms Added &
Addition or modification of domain-specific jargon (field-specific terms, concepts, or processes) not present before and unfamiliar to general audiences. \\
Researcher Names Added &
Inclusion of researcher names not present in the previous revision. \\
\rowcolor{gray!12}
Change in Scientific Narrative &
Change in scientific narrative or perspective---a verb tense, framing, or viewpoint shift---relative to the previous revision. \\
Academic References Added &
Addition of references to published academic papers (identifiers such as DOI, PMID, PMC, ISBN, arXiv, or ISSN) not present before. \\
\rowcolor{gray!12}
Academic References Removed &
Removal of such academic-paper references present in the previous revision but no longer present. \\
Wikilink Added &
Addition of a link to another Wikipedia page not present in the previous revision. \\
\rowcolor{gray!12}
Add./Mod.\ of Quantitative Information &
Addition or modification of statistics or numeric results (metrics, percentages, counts, ratios, confidence intervals, $p$-values, or mathematical expressions) not present before. \\
\midrule
Non Scientific Edit &
No scientifically meaningful modification between revisions; assigned alone. \\
\bottomrule
\end{tabularx}
\caption{The \textsc{WikiStar} taxonomy: ten labels capturing scientifically meaningful edit types
and a \textit{Non-Scientific Edit} label. Definitions provided in our prompt
released with our \coderepo.}
\label{tab:taxonomy}
\end{table}

\section{Appendix - Edit Examples}
\label{app:edit-examples}

\begin{table}[H]
\centering
\footnotesize
\setlength{\tabcolsep}{8pt}
\renewcommand{\arraystretch}{1.3}
\begin{tabularx}{\textwidth}{@{}>{\raggedright\arraybackslash}p{0.20\textwidth} X@{}}
\toprule
\textbf{Edit type} & \textbf{Example (\textit{Previous Section Text}\,$\rightarrow$\,\textit{Edited Section Text})} \\
\midrule
New Sci.\ Information &
[\ldots] modulated by external cues, primarily daylight. $\rightarrow$ [\ldots] modulated by
external cues, primarily daylight. \textbf{They allow organisms to anticipate and prepare
for environmental changes.}\newline
{\scriptsize\textcolor{gray}{(\emph{Circadian rhythm}, lead section; revision of 2008-05-06.)}} \\

\rowcolor{gray!12}
Technical Terms Added &
[\ldots] a protruding tongue and an enlarged tongue near the tonsils, a short neck
$\rightarrow$ [\ldots] an enlarged tongue near the tonsils\textbf{, or macroglossia}, a
short neck [\ldots]\newline
{\scriptsize\textcolor{gray}{(\emph{Down syndrome}, ``Characteristics'' section; revision of 2009-07-22.)}} \\

Quantitative Information &
[\ldots] of the thousands of enumerated ports, \emph{about 250} well-known ports are
reserved by convention $\rightarrow$ [\ldots] \textbf{1024} well-known ports are reserved
by convention [\ldots]\newline
{\scriptsize\textcolor{gray}{(\emph{Port (computer networking)}, lead section; revision of 2014-11-30.)}} \\

\rowcolor{gray!12}
Change in Sci.\ Narrative &
\textit{Prev.}: It is \textbf{easier} to obtain high-quality sequence data when the desired DNA is purified [\ldots] from any contaminants.\newline
\textit{Curr.}: It is \textbf{only possible} to obtain high-quality sequence data when the desired DNA is relatively pure, [\ldots] free from other contaminants.\newline
{\scriptsize\textcolor{gray}{(\emph{DNA sequencing}, ``Large-scale sequencing strategies'' section; revision of 2007-02-07.)}} \\
\bottomrule
\end{tabularx}
\caption{Before/after excerpts for selected \textsc{WikiStar} edit types; changed span in
\textbf{bold}, \texttt{[\ldots]} marks elided text.}
\label{tab:edit-examples}
\end{table}

\clearpage
\section{Appendix - Per-Model Results on \textsc{WikiStar-Bench}}
\label{app:perlabel}

Table~\ref{tab:perlabel_allmodels} reports per-label precision, recall, and F1 for
every model we evaluated on the full benchmark, extending the single-model view of
Table~\ref{tab:perlabel_gpt54}. Metrics are computed against the gold labels.

\begin{table}[H]
\centering
\scriptsize
\setlength{\tabcolsep}{3pt}
\begin{tabular}{@{}l | *{7}{c} | *{7}{c} | *{7}{c}@{}}
\toprule
& \multicolumn{7}{|c}{Precision} & \multicolumn{7}{|c}{Recall} & \multicolumn{7}{|c}{F1} \\
\cmidrule(lr){2-8}\cmidrule(lr){9-15}\cmidrule(lr){16-22}
Label
& \rotatebox{90}{GPT-5 mini} & \rotatebox{90}{GPT-5.4} & \rotatebox{90}{GPT-5.1} & \rotatebox{90}{LLaMA3.3} & \rotatebox{90}{o3-mini} & \rotatebox{90}{GPT-4o} & \rotatebox{90}{Qwen3}
& \rotatebox{90}{GPT-5 mini} & \rotatebox{90}{GPT-5.4} & \rotatebox{90}{GPT-5.1} & \rotatebox{90}{LLaMA3.3} & \rotatebox{90}{o3-mini} & \rotatebox{90}{GPT-4o} & \rotatebox{90}{Qwen3}
& \rotatebox{90}{GPT-5 mini} & \rotatebox{90}{GPT-5.4} & \rotatebox{90}{GPT-5.1} & \rotatebox{90}{LLaMA3.3} & \rotatebox{90}{o3-mini} & \rotatebox{90}{GPT-4o} & \rotatebox{90}{Qwen3} \\
\midrule
New Sci.\ Information & \textbf{.93} & .88 & .86 & .82 & .91 & .92 & .87 & \textbf{.93} & .74 & .92 & .65 & .69 & .48 & .83 & \textbf{.93} & .80 & .89 & .72 & .78 & .63 & .85 \\
Sci.\ Information Removed & .81 & \textbf{.84} & .76 & .59 & .81 & .81 & .46 & .83 & \textbf{.92} & .68 & .33 & .68 & .29 & .09 & .82 & \textbf{.87} & .72 & .42 & .74 & .43 & .15 \\
Sci.\ Clarification Added & .69 & .73 & .54 & .68 & .75 & \textbf{.80} & .53 & \textbf{.81} & .72 & .75 & .59 & .39 & .27 & .66 & \textbf{.75} & .72 & .63 & .63 & .51 & .41 & .58 \\
Technical Terms Added & .88 & .80 & .80 & .87 & \textbf{.92} & \textbf{.92} & .66 & .83 & \textbf{.93} & .83 & .62 & .41 & .36 & .84 & \textbf{.86} & \textbf{.86} & .81 & .72 & .57 & .52 & .74 \\
Researcher Names Added & .77 & \textbf{.93} & .79 & .83 & .90 & .87 & .68 & \textbf{.95} & .82 & .91 & .79 & .78 & .61 & .94 & .85 & \textbf{.87} & .84 & .81 & .84 & .72 & .79 \\
Change in Sci.\ Narrative & .41 & \textbf{.54} & .22 & .21 & .34 & .34 & .25 & .74 & .70 & \textbf{.87} & .55 & .57 & .57 & .63 & .52 & \textbf{.61} & .35 & .30 & .43 & .42 & .36 \\
Academic Refs Added & \textbf{.93} & .90 & .69 & .59 & .88 & .79 & .48 & .93 & .86 & .93 & .94 & .88 & .60 & \textbf{.98} & \textbf{.93} & .88 & .79 & .73 & .88 & .68 & .64 \\
Academic Refs Removed & \textbf{.89} & .84 & .75 & .53 & .85 & .65 & .38 & .82 & \textbf{.86} & .59 & .37 & .46 & .31 & .45 & \textbf{.85} & \textbf{.85} & .66 & .44 & .60 & .42 & .41 \\
Wikilink Added & .94 & .95 & .93 & .91 & .96 & \textbf{.97} & .82 & \textbf{.88} & .83 & .71 & .60 & .52 & .24 & .41 & \textbf{.91} & .89 & .81 & .72 & .67 & .39 & .54 \\
Quantitative Information & .80 & .75 & .69 & .77 & \textbf{.91} & .80 & .59 & \textbf{.89} & .86 & .86 & .62 & .62 & .44 & .78 & \textbf{.84} & .80 & .76 & .69 & .74 & .56 & .67 \\
\cmidrule(lr){1-22}
Non-Scientific Edit & \textbf{.94} & .83 & .86 & .81 & .86 & .63 & .77 & .78 & \textbf{.87} & .67 & .59 & .78 & .85 & .43 & \textbf{.85} & \textbf{.85} & .75 & .68 & .82 & .72 & .56 \\
\midrule
Macro avg & .82 & .82 & .72 & .69 & \textbf{.83} & .77 & .59 & \textbf{.85} & .83 & .79 & .60 & .62 & .46 & .64 & \textbf{.83} & .82 & .73 & .62 & .69 & .54 & .57 \\
Micro avg & .86 & .84 & .74 & .75 & \textbf{.87} & .77 & .65 & \textbf{.86} & .83 & .80 & .61 & .62 & .46 & .64 & \textbf{.86} & .83 & .77 & .67 & .72 & .57 & .64 \\
\bottomrule
\end{tabular}
\caption{\textbf{Per-label performance on \textsc{WikiStar-Bench}.} Precision (P),
Recall (R), and F1 for each model; best value per row-block in \textbf{bold}.
\textit{LLaMA3.3} and \textit{Qwen3} denote LLaMA-3.3-70B and Qwen3-80B.
The LLaMA, GPT-4o, and o3-mini results are evaluated on $N{=}1{,}364$, $N{=}1{,}358$, and $N{=}1{,}357$ examples respectively; small per-model differences reflect responses dropped after parsing.}
\label{tab:perlabel_allmodels}
\end{table}

\section{Appendix - Result Comparison Between Prompt Versions}
\label{app:prompt-versions}

\begin{table}[H]
\centering
\setlength{\tabcolsep}{3pt}
\scriptsize
\begin{tabular}{@{}lcc|cc|cc@{}}
\toprule
                              & \multicolumn{2}{c|}{GPT-5-mini} & \multicolumn{2}{c|}{GPT-5.4} & \multicolumn{2}{c}{LLaMA} \\
\cmidrule(lr){2-3}\cmidrule(lr){4-5}\cmidrule(lr){6-7}
Label                         & Naive & Ref. & Naive & Ref. & Naive & Ref. \\
\midrule
New Sci.\ Information       & .72 & \textbf{.93} & .77 & \textbf{.81} & .73 & \textbf{.84} \\
Sci.\ Information Removed   & .69 & \textbf{.82} & .79 & \textbf{.87} & .46 & \textbf{.49} \\
Sci.\ Clarification Added   & .58 & \textbf{.75} & .62 & \textbf{.73} & \textbf{.63} & .62 \\
Technical Terms Added       & .75 & \textbf{.86} & .78 & \textbf{.86} & \textbf{.71} & \textbf{.71} \\
Researcher Names Added      & .58 & \textbf{.85} & .71 & \textbf{.87} & .73 & \textbf{.81} \\
Change in Sci.\ Narrative   & .40 & \textbf{.52} & .32 & \textbf{.61} & \textbf{.29} & .28 \\
Academic References Added   & .63 & \textbf{.93} & .64 & \textbf{.88} & .63 & \textbf{.72} \\
Academic References Removed & .70 & \textbf{.85} & .72 & \textbf{.85} & .46 & \textbf{.50} \\
Wikilink Added              & .86 & \textbf{.91} & .91 & .89          & \textbf{.81} & .74 \\
Quantitative Information   & .78 & \textbf{.84} & .69 & \textbf{.80} & \textbf{.76} & .73 \\
\cmidrule(lr){1-7}
Non Scientific Edit          & .56 & \textbf{.85} & .63 & \textbf{.85} & \textbf{.63} & \textbf{.63} \\
\midrule
Macro F1                      & .66 & \textbf{.83} & .69 & \textbf{.82} & .62 & \textbf{.64} \\
Micro F1                      & .68 & \textbf{.86} & .72 & \textbf{.83} & .67 & \textbf{.70} \\
\bottomrule
\end{tabular}
\caption{Per-label F1 for the \textit{naive} prompt (a single-sentence definition per label) versus our
\textit{refined} prompt (explicit inclusion/exclusion rules, per-label examples, and an
up-front scientific/non-scientific decision), across three backbone models. \textbf{Bold}
marks each refined score that improves on its naive counterpart. \textit{LLaMA} denotes LLaMA-3.3-70B.}
\label{tab:naive-vs-refined}
\end{table}